\documentclass{article}

\usepackage{arxiv}

\usepackage[utf8]{inputenc} 
\usepackage[T1]{fontenc}    
\usepackage{hyperref}       
\usepackage{url}            
\usepackage{booktabs}       
\usepackage{amsfonts}       
\usepackage{nicefrac}       
\usepackage{microtype}      
\usepackage{lipsum}
\usepackage{graphicx}
\graphicspath{ {./images/} }

\usepackage[table]{xcolor}

\title{A Rigorous Machine Learning Analysis Pipeline for Biomedical Binary Classification: Application in Pancreatic Cancer Nested Case-control Studies with Implications for Bias Assessments}

\author{
Ryan J. Urbanowicz \\
  Institute for Biomedical Informatics\\
  University of Pennsylvania\\
  Philadelphia, PA, 19104 \\
  \texttt{ryanurb@upenn.edu} \\
   \And
Pranshu Suri \\
  Institute for Biomedical Informatics\\
  University of Pennsylvania\\
  Philadelphia, PA, 19104 \\
  \texttt{psuri@seas.upenn.edu} \\
  \And
 Yuhan Cui \\
  Institute for Biomedical Informatics\\
  University of Pennsylvania\\
  Philadelphia, PA, 19104 \\
  \texttt{yuhan.cui@upenn.edu} \\
  \And
 Jason H. Moore \\
  Institute for Biomedical Informatics\\
  University of Pennsylvania\\
  Philadelphia, PA, 19104 \\
  \texttt{jhmoore@upenn.edu} \\
  \And
 Karen Ruth \\
  Biostatistics and Bioinformatics Facility\\
  Fox Chase Cancer Center\\
  Philadelphia, PA, USA \\
  \texttt{karen.ruth@fccc.edu} \\
    \And
Rachael Stolzenberg-Solomon \\
  Division of Cancer Epidemiology and Genetics\\
  National Cancer Institute\\
  Shady Grove, MD, USA \\
  \texttt{rachael.solomon@nih.gov} \\
    \And
Shannon M. Lynch \\
  Cancer Prevention and Control\\
  Fox Chase Cancer Center\\
  Philadelphia, PA, USA \\
  \texttt{shannon.lynch@fccc.edu} \\
}

\begin{document}
\maketitle
\begin{abstract}
Machine learning (ML) offers a collection of powerful approaches for detecting and modeling associations, often applied to data having a large number of features and/or complex associations. Currently, there are many tools to facilitate implementing custom ML analyses (e.g. scikit-learn). Interest is also increasing in automated ML packages, which can make it easier for non-experts to apply ML and have the potential to improve model performance. ML permeates most subfields of biomedical research with varying levels of rigor and correct usage. Tremendous opportunities offered by ML are frequently offset by the challenge of assembling comprehensive analysis pipelines, and the ease of ML misuse. In this work we have laid out and assembled a complete, rigorous ML analysis pipeline focused on binary classification (i.e. case/control prediction), and applied this pipeline to both simulated and real world data. At a high level, this 'automated' but customizable pipeline includes a) exploratory analysis, b) data cleaning and transformation, c) feature selection, d) model training with 9 established ML algorithms, each with hyperparameter optimization, and e) thorough evaluation, including appropriate metrics, statistical analyses, and novel visualizations. This pipeline organizes the many subtle complexities of ML pipeline assembly to illustrate best practices to avoid bias and ensure reproducibility. Additionally, this pipeline is the first to compare established ML algorithms to 'ExSTraCS', a rule-based ML algorithm with the unique capability of interpretably modeling heterogeneous patterns of association. While designed to be widely applicable we apply this pipeline to an epidemiological investigation of established and newly identified risk factors for pancreatic cancer to evaluate how different sources of bias might be handled by ML algorithms.
\end{abstract}


\section{Introduction and Background}
Machine learning (ML) has become a cornerstone of modern data science with applications in countless research domains including biomedical informatics, a field synonymous with noisy, complex, heterogeneous, and often large-scale data (i.e. 'big-data') \cite{thornton2004genetics,luo2016big,rauschert2020machine}. Surging interest in ML stems from it's potential to train models that can be applied to make predictions as well as discover complex multivariate associations within increasingly large feature spaces (e.g. 'omics' data as well as integrated multi-omics data) \cite{diao2018biomedical}. Increased access to powerful computing resources has fueled the practicality of these endeavors \cite{elsebakhi2015large}. As a result, a wealth of ML tools, packages, and other resources have been developed to facilitate implementation of custom ML analyses. One popular and accessible example includes the scikit-learn library built with the Python programming language \cite{pedregosa2011scikit}. Packages such as scikit-learn focus on facilitating the use of individual elements of an ML analysis pipeline (e.g. cross validation, feature selection, ML modeling), however 'how' these elements are brought together is generally left up to the practitioner. Unfortunately, there are no absolute standards or guidelines on how to construct an ML analysis pipeline, but rather more of a community knowledge pool of 'potential pitfalls', 'best practices', and varying degrees of 'rigor' when comparing one analysis pipeline to the next \cite{smialowski2010pitfalls,luo2016guidelines,garreta2017scikit,vieira2020step,uccar2020effect}. This knowledge pool is vast, constantly evolving, often presented from an application-specific perspective, and is distributed over research papers, textbooks, an enormous diversity of web-based documentation, and educational resources. Ultimately, ML constitutes a powerful collection of methods that have become relatively easy to apply at the consequence of being extremely easy to misuse, especially for those who do not specialize in their application or development \cite{hoffman2013use,yapo2018ethical}. 

Most ML research tends to focus on improving or adapting individual methods for a given data type, task, or domain of application. Surprisingly few works have focused on how to effectively and appropriately assemble an ML pipeline in its entirety or provide accessible, easy to use examples of how to do so. For those coming from various biomedical specialties it can be daunting to know where to start with respect to conducting ML analyses and (while their numbers are increasing) there are still a limited number of experts with the appropriate interdisciplinary skills needed to develop an analysis pipeline for a given biomedical data task. As a result, over the last few years, automated ML, or 'auto-ML' packages have taken the spotlight \cite{waring2020automated}. Auto-ML approaches currently tend to fall under one of four categories: (1) automated data cleaning \cite{lucas2019clarite}, (2) automated feature engineering \cite{la2020learning}, (3) automated hyperparameter optimization \cite{akiba2019optuna}, and (4) automated pipeline assembly \cite{olson2019tpot,feurer2020auto,ledell2020h2o}. Pipeline assembly frameworks, in particular, effectively seek to 'stand-in-for' ML experts with the automated convenience of artificial intelligence, such that the non-experts can more easily apply ML. Auto-ML promises greater accessibility and a reduction in the tedium of developing and optimizing an ML analysis. However, these systems can be extremely computationally expensive, and ultimately may not yield the best results as they are ultimately limited to explore within the scope of their programming \cite{truong2019towards,fabris2019analysing}. The use of auto-ML also arguably runs the risk of potentially creating new opportunities for ML misuse when practitioners are not forced to think through and design their analyses when dealing with potential biases introduced by either the study design/data source, or the ML analyses conducted. Further, in focusing solely on optimizing the final model, auto-ML, in itself, does little to further our understanding of ML modeling algorithm performance in different contexts towards improving analysis pipeline optimization and best practices as well as auto-ML strategies of the future. 

\subsection{ML Pipeline Overview} 
One major goal of this paper is to develop, apply, and disseminate an accessible but rigorous ML analysis pipeline focused on binary (e.g. case/control) biomedical classification problems in both simulated and real world data. Through this work, we seek to coalesce and explain essential ML best practices for avoiding pitfalls and offer a complete binary class ML pipeline with many convenient features that can be applied 'as is' or used as a starting point for further customization and improvement. At a high level this pipeline includes: a) exploratory analysis, b) data cleaning (e.g. missing data) and transformation, c) feature selection, d) application of 9 established ML modeling approaches, each with hyperparameter optimization, and e) evaluation, including appropriate metrics, statistical analyses, and novel visualizations. This pipeline is best suited for a) comparing the performance of different algorithms for ML modeling on a given dataset, b) evaluating the importance of potentially relevant features in the dataset; an essential part of model interpretation, c) training a predictive model for downstream application, and/or d) running the pipeline on different datasets to compare achievable performance between them. This pipeline is organized within a well documented Jupyter notebook and coded in Python using well known packages such as pandas, scikit-learn, and scipy.

 Each individual method utilized in this pipeline has been previously published, thus this work focuses on what 'methods to include', and 'how they should be ordered or utilized' as a proposed basic standard of biomedical ML analysis rigor. Novel focuses of this pipeline include: a) utilizing a sequence of methods that preserve the potential to detect and model complex associations, i.e. epistasis \cite{moore2016epistasis} and/or genetic heterogeneity \cite{urbanowicz2013role}, b) adopting collective feature selection \cite{verma2018collective}, c) applying a broader selection (i.e. more than 1 to 3) of well-established and cutting-edge ML modeling algorithms, that ideally have complementary strengths and weaknesses providing multiple 'perspectives' on modeling associations, d) identifying important hyperparameters for optimizing each ML algorithm with suggested value options/ranges for conducting a comprehensive hyperparameter sweep e) performing the first comparison of our cutting edge, interpretable rule-based machine learning algorithm (i.e. ExSTraCS \cite{urbanowicz2015exstracs}) to other well known ML algorithms, f) automatically reporting a wide variety of appropriate evaluation metrics with statistical comparisons and relevant visualizations, g) proposing a novel composite bar plot to illustrate feature importance consensus across different ML algorithms, and h) incorporating a number of convenient pipeline features including; automated generation of figures, the use of 'Optuna' for automated Bayesian hyperparameter optimization \cite{akiba2019optuna}, and the use of Python's 'pickle' module applied to save/store all ML models for downstream replication analysis or application to make predictions on subjects with an unknown outcome. We also offer suggestions on how this pipeline can be further improved or adapted to other target applications and data challenges. This work does not: a) provide a comprehensive comparison of all ML best practices, b) suggest this pipeline is the 'best' or 'only' valid way to assemble an ML pipeline, c) focus on largely dataset-specific elements of an ML pipeline such as feature engineering, algorithm-specific data transformations, or the handling of outliers, or d) replace the need for careful data collection and the design of a study.  In fact, utilizing the standardized statistical assessments and data visualizations of this proposed pipeline, we are able to demonstrate how bias introduced by study design within a data source could impact both feature and ML selection, and subsequent study findings.   

\subsection{Applying ML to an Epidemiologic Cohort Study of Pancreatic Cancer for Bias Assessment} 
The second major goal of this paper is to apply this ML pipeline to a real-world epidemiological investigation of pancreatic cancer, ranked the 14th most common cancer and the 7th highest cause of cancer mortality globally \cite{mcguigan2018pancreatic}. While it is now common to apply ML methods to -omics and electronic medical records (EMR) data within the fields of bioinformatics and biomedical informatics \cite{larranaga2006machine,kleiman2019machine}, there are comparatively fewer ML applications using prospective epidemiologic cohort data where results target cancer prevention and early detection \cite{gail1989statistical}.  In cohort studies, a multitude of exposures related to demographic (race/ethnicity, gender), lifestyle behaviors (smoking habits, diet, physical activity), and medical history (diabetes, excess weight, etc.) are collected using questionnaires from thousands of healthy participants at study entry \cite{zhu2013prostate,bao2016origin}. Participants are then followed over time for disease development, allowing for the establishment of temporal relationships between measured exposures and disease (i.e. certainty that exposures occurred prior to disease development; a current limitation of often cross-sectionally collected EMR data) \cite{breslow1980statistical,rothmangs}. Existing population-based cohort studies often also collect biospecimens that are used to measure genetic and other biomarkers, which add to existing survey data \cite{zhu2013prostate,bao2016origin}. However, rarely have all available survey, biomarker, and genetic features been utilized in combination, given epidemiologic investigations traditionally a) focus on associations between a single exposure with an outcome, and b) use reductionist approaches from standard regression methods that do not empirically consider how multiple risk factors interact or apply heterogeneously to influence cancer outcomes \cite{lynch2016call}.

While ML methods can facilitate full utilization of cohort data, a lack of understanding of “what” ML algorithms are doing “behind the scenes” has limited their application in population-based cohorts \cite{mooney2015epidemiology}. Contributing to this “black box” perception is a weak understanding of how ML methods handle bias or systematic error in research methods, which can influence disease associations and prediction. For instance, nested case-control studies are often conducted within cohort studies to capitalize on the prospective, temporal nature of the data \cite{gail1989statistical}.  In nested case-control studies, selection of cases and controls (i.e. participants with and without disease from the full study cohort) are often conducted based on data availability, which can introduce bias \cite{rothman1998gs}. For example, selecting only participants from a cohort study with available biospecimen data might lead to an analysis sample that is not representative of the total sample (e.g. selection bias), and/or could introduce confounding, which occurs when  true effects of a variable are inaccurate due to unequal distributions of a related variable by case/control status \cite{rothman1998gs}. To address the impact of sample selection on study external validity (i.e. generalizability), replication of findings across multiple population samples is commonly recommended. To improve internal study validity, statistical adjustments for biases, such as confounding, are traditionally employed. However, in a ML setting, it is likely that comparisons of multiple methods and the identification and comparisons of feature importance within the same study would be needed to inform both study validity, as well as address the "black box" nature of ML selection, and subsequent study interpretations. However, this remains an understudied area, particularly in epidemiologic investigations. 

To that end, this study applies the proposed ML pipeline, with 9 distinct ML algorithms, to three nested case-control data subsets derived from the Prostate, Lung, Colorectal, and Ovarian Cancer (PLCO) Trial cohort population, focusing on pancreatic cancer as the diagnostic outcome and features comprised of primarily survey variables that measures both established and potential risk factors for pancreatic cancer. Each subset is characterized by different bias considerations related to case/control selection and confounding (Section \ref{panc_data}). We evaluate ML model predictive performance and feature importance across the three nested case-control subsets in order to a) compare ML algorithm performance, b) determine whether ML algorithms identify known informative features  (i.e. replicate findings related to established risk factors) and/or to identify new or suspected risk factors as informative (namely dietary variables), c) assess the influence of bias on ML model performance and ML algorithm selection. While the application of the ML pipeline will likely provide insight into relevant risk factors for pancreatic cancer, as well as potential interactions, the primary purpose of this application is not for discovery, but to a) demonstrate the application of our pipeline to real world data, b) evaluate the ability of the pipeline to identify established risk factors as informative, c) inform case/control selection, ML selection, and bias adjustments in future combined analyses that will include additional genetic and biomarker data.

The following sections will explain our proposed binary classification ML analysis pipeline, including design justification, integration of ML pipeline best practices, and useful automated features. Then we will detail the pancreatic cancer datasets and evaluation of ML bias. This will be followed by analysis results, interpretation, discussion conclusions, and future work. 

\section{Methods} \label{methods}
This section details: a) our proposed ML pipeline for binary classification and b) the target pancreatic cancer datasets and considerations for evaluating ML bias. In designing this pipeline we sought to give each of the ML modeling algorithms the opportunity to perform their best, as well as to give the overall analyses the ability to detect not only simple univariate effects and multivariate additive effects, but complex associations with outcome including feature interactions (e.g. epistasis) as well as heterogeneous feature associations  (e.g. genetic heterogeneity or phenocopy), where different unspecified subgroups of instances in the data may have distinct underlying feature associations with outcome. While most methods are proficient at detecting simpler associations, more sophisticated ML algorithms are often required to detect complex interactions. To date, only rule-based ML algorithms such as ExSTraCS \cite{urbanowicz2015exstracs} have been explicitly demonstrated to perform well in detecting heterogeneous feature associations. This pipeline offers a framework to explore whether, or to what degree, other well-known ML algorithms also have this capability.

\subsection{Introducing the Binary Class ML Pipeline using Simulated Data} \label{mlpipe}

\begin{figure}[h]
\centering\includegraphics[width=1.0\linewidth]{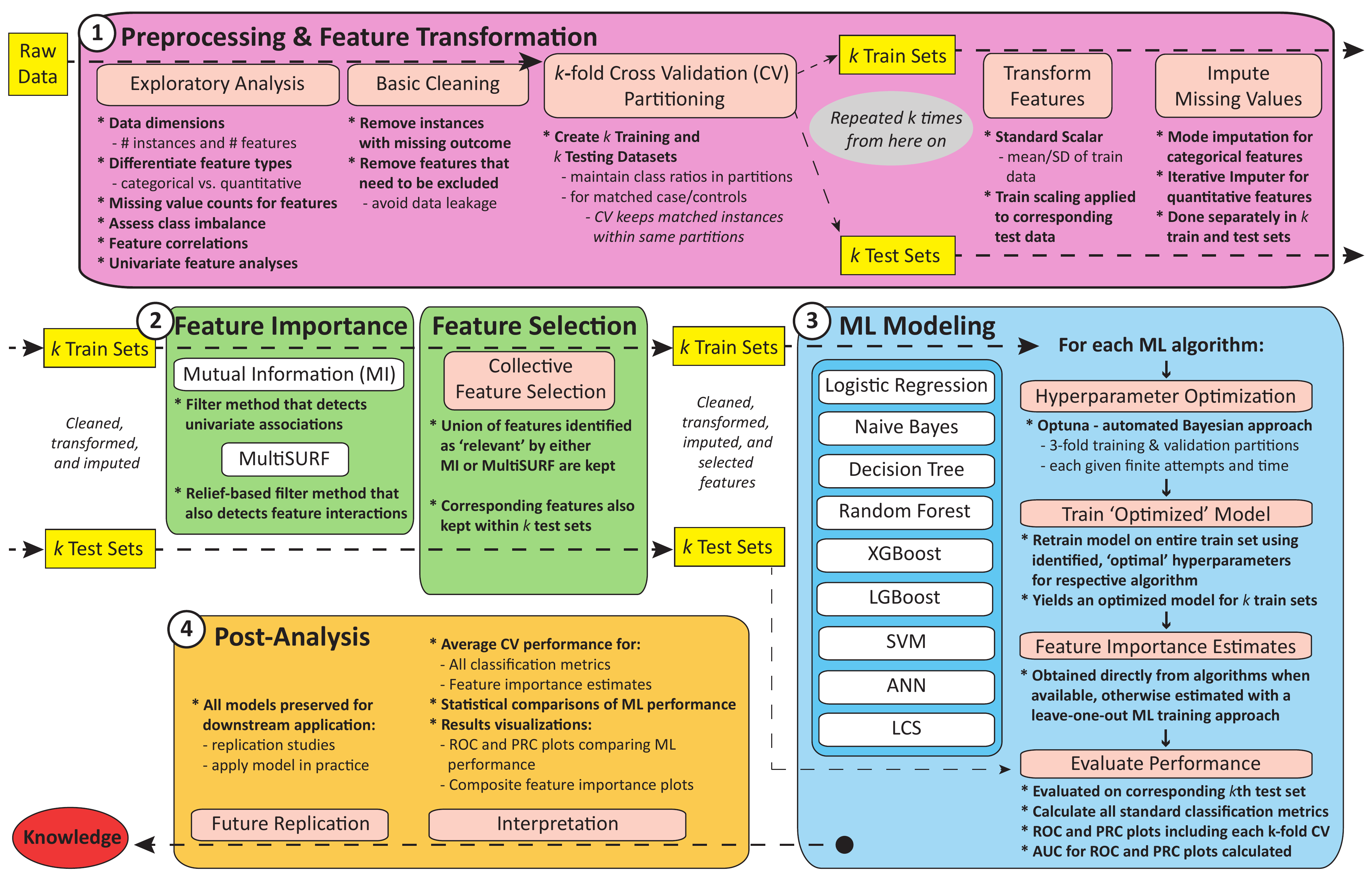}
\caption{Schematic overview of our proposed ML analysis pipeline. Starting from a single 'raw' dataset the pipeline proceeds through 4 main stages towards the output of 'knowledge'. Arrows indicate pipeline flow, i.e. the specific sequence of steps. The raw dataset and subsequent training and testing datasets are presented as yellow boxes. More specific pipeline elements are highlighted in light pink boxes. Specific algorithms are highlighted in white boxes. The arrows leaving box 1 correspond to the arrows entering box 2 below it.}
\label{ml_pipe}
\end{figure}

As summarized in Figure \ref{ml_pipe} our proposed ML pipeline is compartmentalized into 4 main stages: 1) data preprocessing and feature transformation, 2) feature importance and feature selection, 3) ML modeling, and 4) post-analysis. Subsections \ref{prepro} - \ref{post} below detail these stages. This pipeline has been implemented as an accessible Python-based Jupyter notebook for biomedical researchers that may be relatively new to ML or Python programming. It can be run easily on new datasets with minor notebook changes, or it can be modified to suit the specific needs of a given dataset analysis. The notebook and required Python project files can be downloaded at the link below with installation instructions and software/package requirements included.

\begin{center}
\texttt{https://github.com/UrbsLab/ExSTraCS$\_$ML$\_$Pipeline$\_$Binary$\_$Notebook}
\end{center}

This pipeline can be applied to a given dataset by ensuring the following minimum values have been properly specified in the notebook: a) dataset name, b) class label, c) instance ID label (if present) and d) the list of any feature labels that should be excluded from the analysis. Users have the option to adjust a wide range of additional pipeline settings. In order to test this pipeline as well as illustrate it's functionality the description that follows is paired with results applying it to a simulated single nucleotide polymorphism (SNP) dataset generated using the GAMETES software \cite{urbanowicz2012gametes}. This simulated dataset includes 20 SNP features and 1600 instances with no missing values. Four of the features (M0P0, M0P1, M1P0, and M1P1) are predictive of a balanced binary outcome, simulated with minor allele frequencies of 0.2 and heritability of 0.4. Remaining features were randomly generated with minor allele frequencies between 0.05 and 0.5. The four predictive features have been simulated such that there are no univariate associations, but the pairs (M0PO and M0P1) and (M1P0 and M1P1) model purely epistatic interactions each of which is heterogeneously associated with outcome within two respective halves of the training instances \cite{urbanowicz2018benchmarking}. All pipeline results in this paper, including this simulated dataset example, are available at the link below: 

\begin{center}
\texttt{https://github.com/UrbsLab/Pancreatic$\_$Cancer$\_$ML$\_$Notebook$\_$Analysis}
\end{center}

\subsubsection{Preprocessing and Feature Transformation} \label{prepro}
Given a 'raw' dataset it is established best-practice to conduct an \emph{exploratory analysis}. This is of particular import for investigators who may not have collected or worked with this data previously. The purpose of an exploratory analysis is primarily to understand the data and it's characteristics in order to properly guide downstream analysis, as well as to identify potential problems/errors in the data that will need to be addressed. In this pipeline, exploratory analysis is intertwined with a simple, universal \emph{data cleaning} that a) removes any instances with a missing outcome value (i.e. class), which hold no value in supervised learning, and b) removes any dataset columns that should be excluded from the analysis, which may be important for preventing data leakage (i.e. giving the ML algorithm information during training it should not have access to) \cite{smialowski2010pitfalls}. Other, dataset-specific data cleaning steps that are not included, but may also be appropriate at this stage include outlier removal \cite{lucas2019clarite}, re-encoding features (e.g. one-hot-encoding of categorical features) \cite{cerda2018similarity}, or the removal of data rows or columns with missing values (if imputation is undesirable)\cite{batista2003analysis}. This pipeline's exploratory analysis reports: a) the number of instances and features, b) missing value counts by feature, c) the class counts, to evaluate potential class imbalance, d) the number variables that will be treated as categorical vs. quantitative, e) a feature correlation analysis, to identify potentially redundant or highly correlated features, and f) appropriate statistical univariate analyses between individual features and outcome where, automatically, categorical features are analyzed via Chi Square test and visualized with a bar plot, and quantitative features are analyzed via Mann-Whitney test an visualized with a box plot. In our simulated data example, no features were identified as significant after univariate analyses, as expected.

Class imbalance is particularly important to identify, since it can impact ML modeling performance \cite{luque2019impact}. This pipeline leaves any class imbalance, 'as is', and ensures the inclusion of downstream evaluation metrics that take class imbalance into account (i.e. using balanced accuracy and precision-recall curves). \emph{Balanced accuracy} is the average of sensitivity and specificity \cite{urbanowicz2015exstracs}. Common alternatives for addressing class imbalance include either over-sampling the minority class (which runs the risk of adding sampling bias), or under-sampling the majority class (which reduces sample size and power to detect signal) \cite{japkowicz2000class}. 

With respect to feature correlation, this pipeline does not apply any strategy to remove correlated features, since the removal of highly, but not perfectly, correlated features has the potential to eliminate relevant signal, in particular with respect to feature interactions \cite{yu2004efficient}. While it may be reasonable to remove highly correlated, and thus potentially redundant features in certain contexts, ultimately it is only truly safe to remove all but one of a pair or set of 'perfectly correlated' features. Due to the context specific nature of this task, our pipeline does not automate redundant feature removal, and assumes that no perfectly correlated features are present. 

The next step separates the cleaned dataset into \emph{k} training and testing datasets using \emph{k-fold cross validation} (CV) partitioning \cite{purushotham2011evaluation}. Three strategies for CV partitioning have been implemented: a) random, b) stratified, and c) matched. Stratified CV is employed to ensure that class ratios are maintained within each partition. Matched CV similarly maintains class ratios, but additionally ensures that cases and controls that had been previously 'matched' by covariate value(s) to avoid epidemiological confounding, are kept within the same partitions (Section \ref{panc_data}). Notably, from this point on all steps in the pipeline are repeated independently for each of the \emph{k} sets. It is critical that all other elements of this pipeline take place after CV partitioning to avoid various biases from being introduced. Specifically, instances in respective testing sets can, in no way, be applied such that they might impact model training. This includes feature transformation methods that rely on characteristics of the data (e.g. mean or standard deviation), missing data imputation, feature selection, and of course ML modeling itself. In this entire study a \emph{k} of 10 was used. 

Next, this pipeline applies a standard scalar approach as a \emph{feature transformation} and normalization step \cite{singh2019investigating}. Standard scaling operates on each feature independently, removing the mean and scaling to unit variance. This step is important when features in the dataset have different ranges which can impact the performance and interpretability of some ML modeling algorithms (e.g. logistic regression, support vector machines, and artificial neural networks \cite{singh2019investigating}). Importantly, the feature means and standard deviations for standard scaling are calculated from the training sets to preserve the unseen nature of hold out testing data.  Once calculated, this pipeline preserves these transformations as Python 'pickle' files. Pickling in Python is a convenient way to serialize and de-serialize objects (such as data, code objects, or trained ML models) for future use. Saved training set transformations are applied to their corresponding testing sets, and can be applied in the future to replication data, or when applying trained ML models to make predictions in real world applications.

The last element of preprocessing calls for the \emph{imputation} of missing data values. Imputation is the process of replacing any missing data with substituted values. Imputation is effectively the process of manufacturing data by means of an 'educated guess'. As such, there is ample opportunity to undesirably bias analyses \cite{white2010avoiding}. While some ML algorithms and implementations can operate in the presence of missing values, our ML pipeline relies on various scikit-learn packages that, by default, cannot. As such, we view imputation here as a 'necessary evil' that will afford us the opportunity to more fairly compare a range of ML modeling algorithms. This notebook first applies 'mode' imputation to categorical features, i.e. the most common feature value will be substituted for missing values, and then applies a more sophisticated iterative imputation approach \cite{buuren2011mice} to quantitative features. This ensures that no additional categorical states are added by the imputation process. Unlike mode imputation, iterative imputing considers the context all features in the dataset. Imputation is conducted separately in training and testing datasets, so that information from the testing data is excluded from downstream model training.

\subsubsection{Feature Processing} \label{featpro}
The next stage of this pipeline serves two roles: a) evaluating \emph{feature importance} outside of a particular ML modeling algorithm and b) \emph{feature selection} (FS), seeking to remove irrelevant features from the data and reduce the size of the feature space for downstream modeling. FS is an important precursor to ML modeling since the removal of irrelevant features will reduce opportunity for overfitting and is likely to yield better performing and more interpretable models. This is particularly important when analyzing datasets with very large feature spaces (i.e. big data). While FS can take place in parallel with ML modeling (i.e. wrapper-based or embedded FS \cite{urbanowicz2018relief}), this can be computationally expensive, difficult to automate, and can identify feature subsets that are biased by the modeling algorithms being applied. Therefore this pipeline focuses on utilizing filter-based FS methods, which are typically fast and can be combined with any downstream ML method. Unfortunately, most filter-based FS methods are insensitive to complex feature interactions. For example, when conducting a genome-wide association study, it is common to reduce the feature space by first removing all features without a significant univariate association with outcome, prior to downstream analysis. This runs the risk of removing features that may informative only in combination with other features. Ultimately we are concerned with detecting both simple and potentially complex associations. 

\begin{figure}[h]
\centering\includegraphics[width=0.9\linewidth]{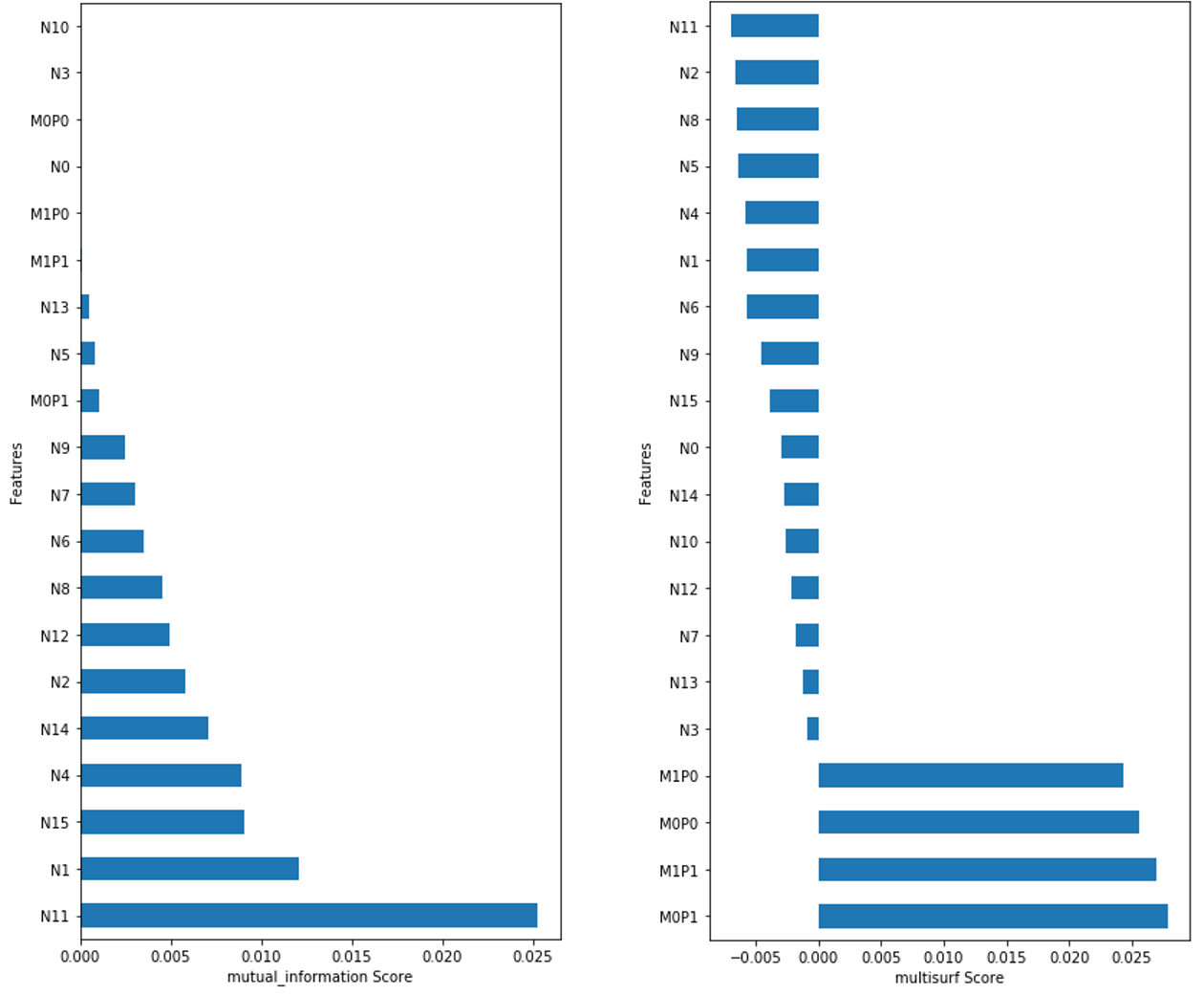}
\caption{Average feature importance results across 10-fold CV on simulated SNP data. The left barplot gives MI scores and the right barplot gives MultiSURF scores. Simulated predictive feature names start with an 'M' while non-predictive features start with an 'N'. Note how MultiSURF perfectly prioritizes the four features simulated with epistatic and heterogenous patterns of association while MI does not.}
\label{sim_fi}
\end{figure}

\emph{Collective feature selection} is an FS 'ensemble' approach that calls for the use of more than one FS methodology in making the determination as to whether a given feature should be retained or removed \cite{verma2018collective}. If either method finds evidence that a feature may be informative, that feature is retained, otherwise it is removed. Here we implement two FS algorithms as part of a collective FS scheme, i.e. mutual information (MI) \cite{peng_feature_2005} and MultiSURF \cite{urbanowicz2018benchmarking} (a Relief-based FS algorithm).  MI is proficient at evaluating univariate associations between a feature and outcome, while MultiSURF has been demonstrated as being sensitive to not only univariate associations but heterogeneous associations and both 2 and 3 way feature interactions (even in the absence of univariate associations). Both algorithms yield feature importance scores for every feature where a score larger than $0$ is considered to be potentially informative (and the feature is retained). Figure \ref{sim_fi} presents average feature importance scores for both algorithms on our simulated example. This example highlights the value of collective feature selection, given that had we only used MI, three of our predictive features (M0P0, M1P0, and M1P1) might have been eliminated from consideration during feature selection. 

This pipeline includes a setting allowing the user to specify a maximum number of features to be preserved prior to ML modeling (set to 50 here). The datasets investigated in this paper all had fewer than 50 features, so it had no impact on this study, but may have a significant impact on other data analyses. It may be useful to extend the variety of FS methods implemented in this pipeline in the future, and further explore how to best design a collective FS approach. Additionally, it has been previously demonstrated that when there are 10,000 features or more, MultSURF can lose power to detect pure interactions. However, Relief wrapper methods such as TuRF \cite{moore2007tuning} and VLSRelief \cite{eppstein2008very} have been proposed that can dramatically boost power in these circumstances. Furthermore, MultiSURF scales quadratically with the number of training instances, therefore this notebook includes a user parameter to control the maximum number of randomly selected instances used by MultiSURF during scoring (set to 2000 by default to avoid very large run-times). 

FS takes place independently for each of the \emph{k} training sets which means that it is possible for each set to have a different subset of features available. This pipeline is set up to ensure the fidelity of final feature importance summaries as a result of this complication, and outputs ranked feature importance scores and bar plot summaries of top scoring features.

\subsubsection{ML Modeling} \label{MLmodel}
The heart of this pipeline is ML modeling itself. ML refers to a large family of algorithmic methodologies that differ with respect to knowledge representation and inductive learning approaches as well as to the types of data and associations they are most effectively applied to and how interpretable their respective models will be. Many biomedical studies presented as ML analyses only utilize 1-3 different algorithms, most commonly applying either logistic regression (LR) \cite{sikorska2013gwas}, decision trees (DT) \cite{motsinger2010grammatical}, random forests (RF) \cite{qi2012random}, and/or support vector machines (SVM) \cite{mittag2012use}. Deep learning (DL) has become extremely popular in recent years, often mistaken by novices as being equivalent to ML itself. In summary, DL is simply an elaboration of an artificial neural network (ANN), that typically includes more than 4 or 5 'hidden-layers' within the model's architecture \cite{miotto2018deep}. While DL methods have been extremely successful in certain ML applications such as image classification \cite{rawat2017deep}, they are notoriously computationally intensive, difficult to interpret, and demand a large sample size for training \cite{najafabadi2015deep}. It is for these reasons, that while ANN is included in this pipeline, DL is not. In addition to LR, DT, RF, SVM, and ANN, this pipeline also utilizes Naive Bayes (NB) \cite{thakkar2010health}, XGBoost \cite{chen2016xgboost}, LGBoost \cite{ke2017lightgbm}, and our own rule-based ML algorithm called ExSTraCS \cite{urbanowicz2015exstracs}. ExSTraCS is a Learning Classifier System (LCS) algorithm which combines evolutionary learning with a 'piece-wise' knowledge representation comprised of a set of IF:THEN rules \cite{urbanowicz2009learning,urbanowicz2017introduction}. Unlike most ML methods, LCS algorithms learn iteratively from instances in the training data and adapt a population of IF:THEN rules to cover unique parts of the problem space. Notably, DT, RF, XGBoost, and LGBoost are a set of increasingly sophisticated tree-based ML algorithms. XGBoost and LGBoost are currently two of the most popular and successful ML algorithms within the ML research community. We selected this assortment of 9 ML algorithms to strike a balance between variety and computational expense. We expect future work to expand this assortment and believe that ultimately there is merit in identifying a minimal subset of ML algorithms with complementary strengths and weaknesses to apply in future studies. 

ExSTraCS and other related rule-based ML algorithms are not well-known within the ML or biomedical research communities. However, ExSTraCS has been demonstrated to have the ability to interpretably detect, model, and characterize complex associations such as epistasis and genetic heterogeneity. Uniquely, ExSTraCS not only offers a strategy to model heterogeneous patterns of associations but to identify candidate instance subgroups defined by respective predictive features. To date, ExSTraCS has only been compared to other rule-based ML algorithms, thus it is implemented here so that other researchers may easily apply and compare it to other well known methodologies. Based on previous work, we expect ExSTraCS to outperform other algorithms when the data includes heterogeneous associations, and when model interpretability is paramount. Notably in this analysis we evaluate ExSTraCS performance before and after a rule-compaction procedure (i.e. QRF \cite{tan2013rapid}) has been applied to trained models.

\begin{figure}[h]
\centering\includegraphics[width=0.9\linewidth]{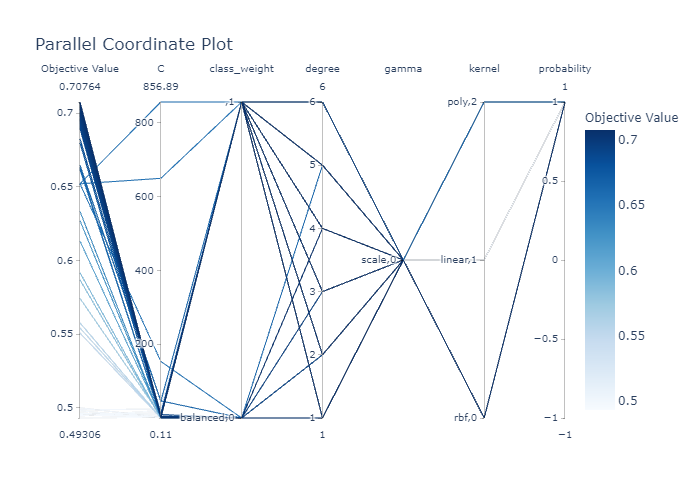}
\vspace{-1cm}
\caption{A hyperparameter sweep visualization output by Optuna for the SVM algorithm on a randomly chosen training partition of the simulated SNP dataset. The objective value in this figure has been set to the evaluation metric, 'balanced accuracy'.}
\label{svm_hyper}
\end{figure}

According to current best practices, the first step in ML modeling is to conduct a \emph{hyperparameter optimization sweep}, a.k.a. 'tuning' \cite{schratz2019hyperparameter}. \emph{Hyperparameters} refer to the run parameters of a given ML algorithm that controls its functioning. Too often, ML algorithms are applied using their 'default' hyperparameter settings. This can lead to unfair ML algorithm comparisons, and a missed opportunity to obtain the best performing model possible. The optimization effectively 'tries out' different hyperparameter settings on a subset of the training instances, ultimately selecting those yielding the best performance to train the final model. The first consideration, is what hyperparameters to explore for each algorithm, as well as the range or selection of hyperparameter values to consider for each. As there is no real consensus regarding this first consideration, we have surveyed a number of online sources, publications, and consulted colleagues in order to select the wide variety of hyperparameters and value ranges incorporated in this pipeline that are suited to binary classification, and detailed in the aforementioned Jupyter notebook. For example, this notebook considers 8 hyperparameters for the optimization of the RF algorithm including 'number of estimators', i.e. the number of trees in the 'forest', with potential values ranging from 10 to 1000. Notebook users can easily adjust the values considered in the hyperparameter optimization of each ML algorithm within the Notebook run parameters. The second consideration, is what approach to take in conducting the hyperparameter sweep. Commonly approaches include a 'grid search' or a 'random search' \cite{uppu2017tuning} which either exhaustively or randomly hyperparameter value combinations. This pipeline adopts a new 'automated' package called 'Optuna', which applies Bayesian optimization of the hyperparameter space given user defined computing constraints \cite{akiba2019optuna}. In this pipeline, Optuna is afforded 100 optimization attempts, applying a nested 3-fold CV procedure and balanced accuracy \cite{urbanowicz2015exstracs} as the evaluation metric. In other words during Optuna optimization the target training data is further partitioned into 3 sub-training sets paired with hold out validation sets.  This way, hyperparameters are selected based on their ability to generalize to unseen data (i.e. their average balanced accuracy on the 3 validation sets). Note, that this optimization does not see that hold out testing data. With Optuna, this pipeline also includes visualizations of each hyperparameter sweep to illustrate how different settings combinations impact performance. Figure \ref{svm_hyper} illustrates one such visualization applying SVM to our simulated data example. Note how the sweep suggests that the 'kernel' hyperparameter performs poorly when set to 'linear', performs better when set to 'poly' and best when set to 'rbf', i.e. radial basis function when applied to this data. 

Optuna optimization is conducted on 7 of the algorithms in this pipeline. Naive Bayes does not have any hyperparameters to optimize, and ExSTraCS is a computationally expensive evolutionary algorithm, making extensive hyperparameter optimization impractical. We have set ExSTraCS to apply reasonable hyperparameters used in previous studies (i.e. a maximum rule population size of 2000, and 200,000 training iterations) \cite{urbanowicz2015exstracs}. This likely puts ExSTraCS at a slight disadvantage compared to other ML algorithms in this pipeline. 

Once 'optimal' hyperarameters have been selected for each of the \emph{k} training datasets and ML algorithms, a 'final' model is trained using the full training set for each algorithm. These models are 'pickled', as described in section \ref{featpro} so that they can be easily utilized in the future. 

Next, this pipeline further examines \emph{feature importance}, now from the perspective of each ML model. These estimate offer useful insights into the high level interpretation of models. Specifically, which features were most important for making accurate predictions. Some algorithm implementations, including DT, RF, and LCS have built in mechanisms to report feature importance estimates after model training, without the need for further computation. In order to obtain feature importance estimates for the other 6 algorithms this pipeline implements a 'leave-one-out' approach were the target algorithm is trained again with the same 'optimal' hyperparameter settings an additional \emph{n} times, on the training data with one of each of the \emph{n} features left out. A relative feature importance estimate is calculated based on how leaving out each feature impacted the model's balanced accuracy. 

\begin{figure}[h]
\centering\includegraphics[width=1.0\linewidth]{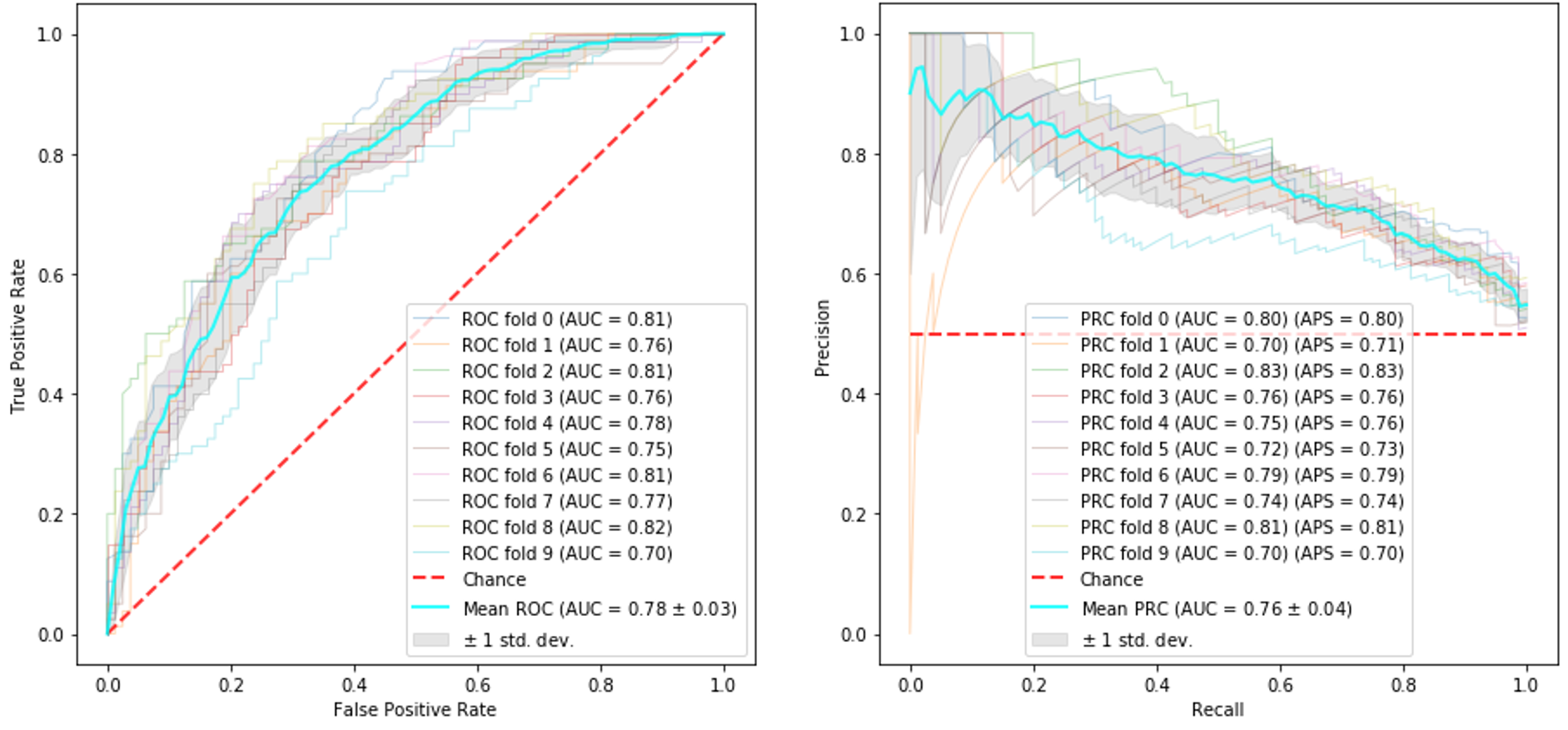}
\caption{Example ROC and PRC plots illustrating LGBoost modeling performance on the simulated SNP data over each 10-fold CV along with mean ROC and PRC, respectively.}
\label{lbg_plots}
\end{figure}

The last step in ML modeling is to evaluate model performance. It is essential to select appropriate evaluation metrics to fit the characteristics and goals of the given analysis. This pipeline ensures that a comprehensive selection of binary classification metrics and visualizations are employed that offer a holistic perspective of model performance. Earlier aspects of this pipeline demanded that a single evaluation metric be selected. We selected \emph{balanced accuracy} as the primary evaluation metric of this notebook, as it equally emphasizes accurate predictions within either class. Balanced accuracy should replace traditional accuracy in the presence of imbalanced data. In addition to accuracy, and balanced accuracy this notebook calculates and reports the following evaluation metrics: a) F1-Score, b) recall, c) specificity, d) precision, e) receiver operating characteristic (ROC) area under the curve (AUC), f) precision-recall curve (PRC) AUC, g) PRC average precision score (APS), h) true positive count, i) true negative count, j) false positive count, and k) false negative count. Additionally for each algorithm, this pipeline outputs an ROC and PRC plot summarizing performance of each of the \emph{k} trained models along side the mean ROC or PRC, respectively. PRC plots are preferable to ROC plots as class imbalance becomes more extreme. This pipeline automatically sets the 'no-skill' lines of PRC plots based on the class ratios in the dataset. Figure \ref{lbg_plots} gives the ROC (left) and PRC (right) plots generated by this pipeline for the LGBoost algorithm.

\subsubsection{Post-Analysis} \label{post}
\begin{figure}[h]
\centering\includegraphics[width=1.0\linewidth]{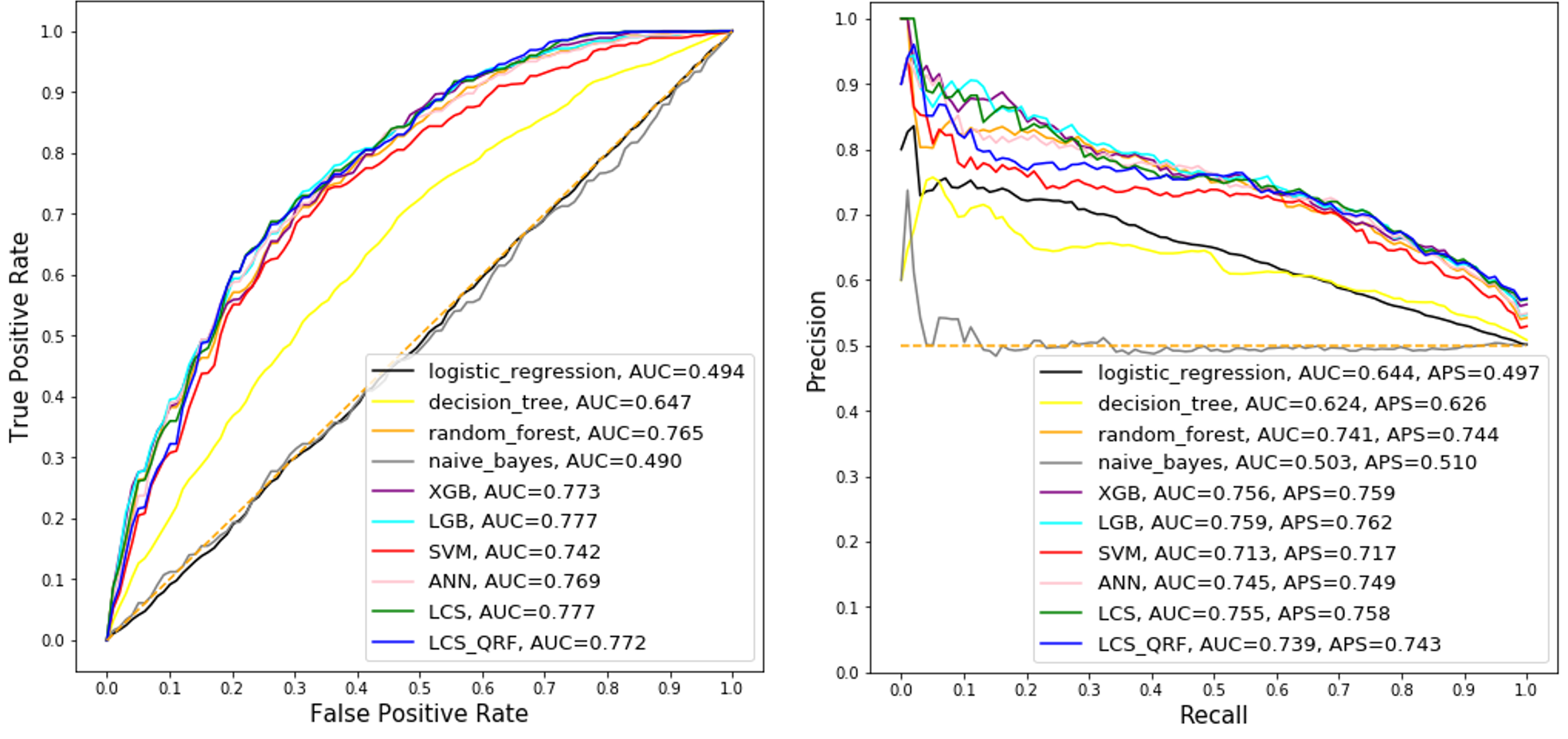}
\caption{ROC and PRC plots comparing average ML algorithm performance on simulated SNP data.}
\label{ml_plots}
\end{figure}

The last stage of this ML analysis pipeline seeks to further summarize evaluation metrics and feature importance estimations, averaged across the \emph{k} partitions to facilitate \emph{interpretation} of the results. First, ROC and PRC plots are generated, that compare average ROC and PRC across all ML algorithms. Figure \ref{ml_plots} gives these plots for our example simulated dataset. Note how both LR and NB fail to detect any signal in the absence of univariate associations and a DT struggles to fully capture the signal in the context of these complex associations. Based on ROC-AUC, LCS and LGB perform best in this data followed by XGB, LCS (following QRF rule compaction), ANN, RF, and SVM performing slightly less well. This should not be taken as an indication that top performing algorithms on this dataset will always perform best, however this confirms that our pipeline has the capability to detect complex associations. Interestingly, this example analysis suggests that, like LCS, XGB and LGB may be effective strategies for modeling associations in the presence of genetic heterogeneity, however, unlike with LCS, neither approach currently offers a path to characterize this heterogeneity or identify candidate heterogeneous patient subgroups \cite{urbanowicz2012analysis,urbanowicz2015exstracs}.

\begin{figure}[h]
\centering\includegraphics[width=1.0\linewidth]{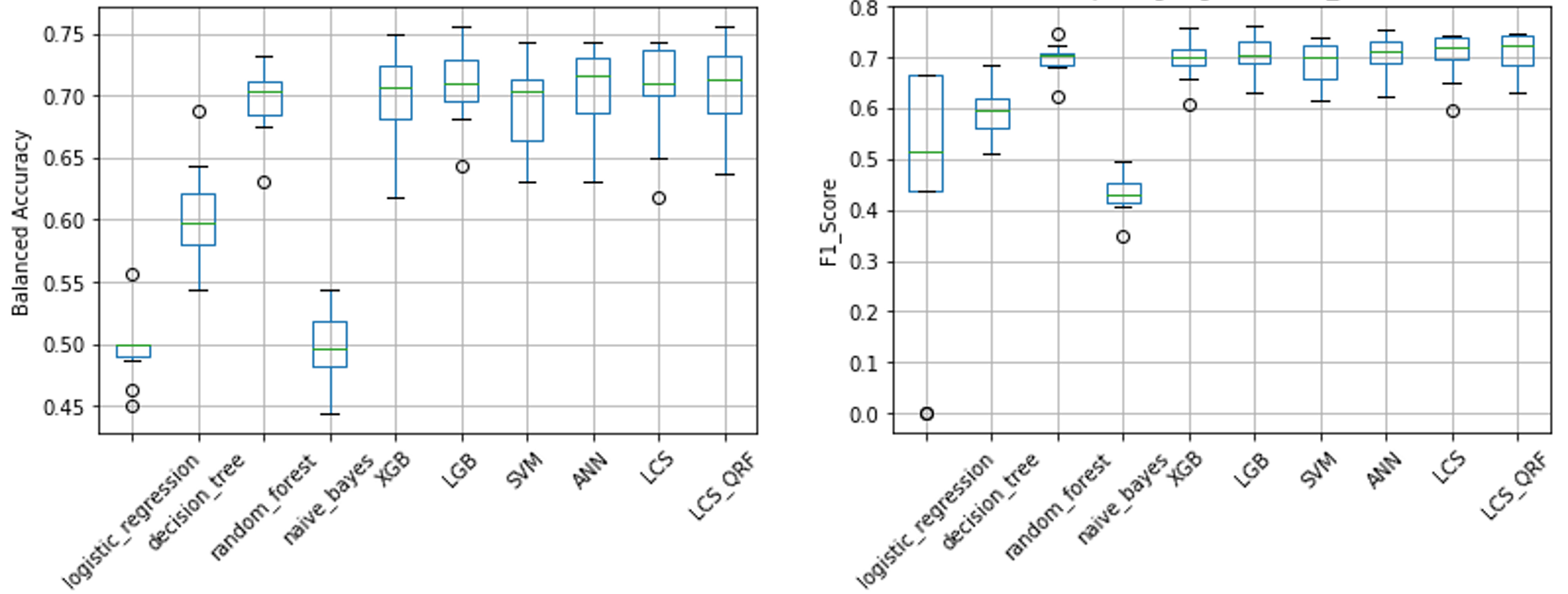}
\caption{Boxplots comparing balanced accuracy (left) and F1-score (right) on simulated SNP data over \emph{k}-fold CV for each ML algorithm. Note the different y-axis ranges.}
\label{metric_boxplots}
\end{figure}

Second, a file summarizing the average and standard deviations of all model performance metrics and every algorithm is output. Third, box plots are generated for each evaluation metric comparing ML algorithm performance over the \emph{k}-fold CV. Figure \ref{metric_boxplots} illustrates these box plots for balanced accuracy and F1 score metrics.

\begin{figure}[h]
\centering\includegraphics[width=1.0\linewidth]{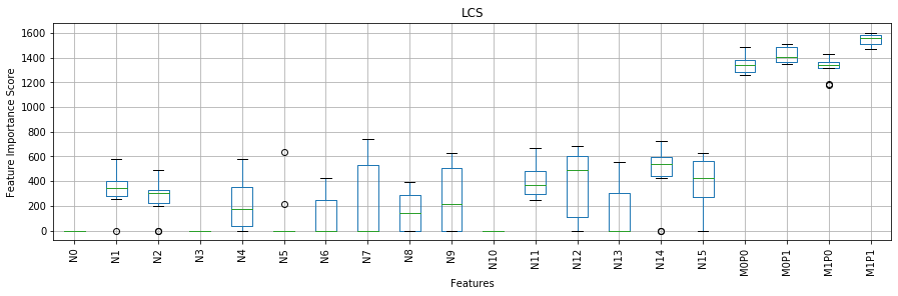}
\caption{Boxplot summarizing LCS feature importance estimates on simulated SNP data across 10-fold CV. Note how the simulated predictive features stand out with the highest scores.}
\label{lcs_fi}
\end{figure}

Fourth, non-parametric statistical analysis comparing average ML performance for each metric is conducted. This includes a Kruskal-Wallis one-way analysis of variance to determine if any ML algorithm performs significantly better or worse for the given evaluation metric. If significant (based on a 0.05 cutoff), then pairwise Mann-Whitney U-tests between ML algorithm pairs are conducted for the given metric. In our example analysis all metrics yielded a significant Kruskal-Wallis test, and follow up pairwise Mann-Whitney U-tests on balanced accuracy confirmed that LR, NB, and DT performed significantly worst than other algorithms, but there was no statistically significant difference in performance between RF, XGB, LGB, SVM, ANN, LCS, or LCS with QRF. Fifth, feature importance box plots are generated for each ML algorithm. Figure \ref{lcs_fi} gives the corresponding box plot for the LCS algorithm, ExSTraCS.

\begin{figure}[h]
\centering\includegraphics[width=1.0\linewidth]{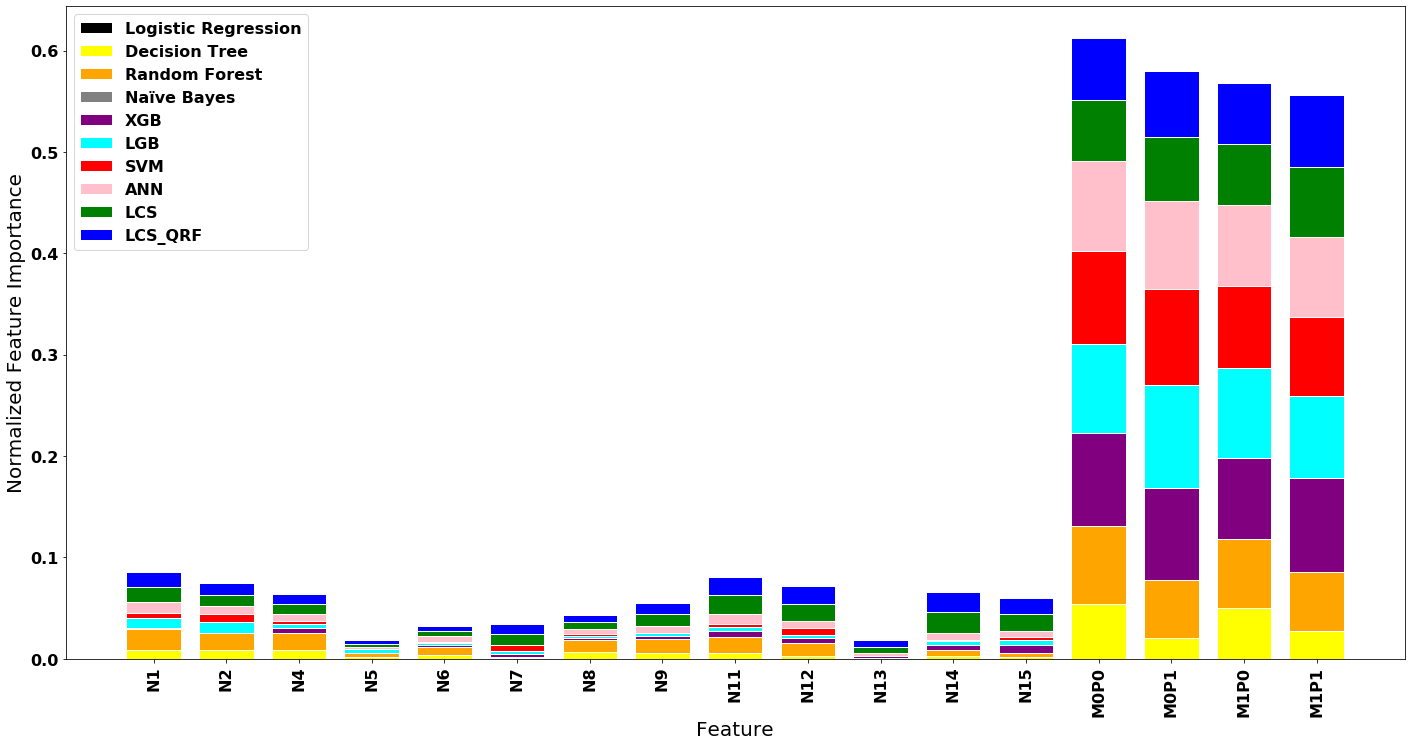}
\caption{Example composite feature importance bar plot (CFIBP) for simulated SNP data applying normalization, sum division, and algorithm performance weighting.}
\label{composite_fi_frac_weight_sim}
\end{figure}

 Lastly, this post-analysis includes the generation of 'composite' feature importance bar plots (CFIBP) that illustrate and summarize feature importance consistency across all ML algorithms. We propose and consider four unique approaches to generating this visualization in uniquely informative ways. The first CFIBP is generated by normalizing the feature importance scores within each algorithm between 0 and 1 and then additively combining the scores for each algorithm  as a single bar per feature. The second CFIBP divides these normalized scores by the sum of scores for the algorithm ensuring that each algorithm has an equal weight in the CFIBP regardless of the range of feature importance scores observed. This is particularly useful if all features receive a similar score by a given ML algorithm, and we do not want that to disproportionately bias the illustration. The third CFIBP instead weights the normalized scores by the average balanced accuracy of the respective algorithm.  This gives an algorithm performance weighted perspective on feature importance. The fourth CFIBP both divides the normalized scores by the sum of scores, and weights them by average balanced accuracy of the respective algorithm. This combination seems to more clearly differentiate features standing out as important across ML algorithms. Figure \ref{composite_fi_frac_weight_sim} gives the fourth CFIBP variant for our example dataset pipeline analysis. 

\subsection{Epidemiologic Study of Pancreatic Cancer to Assess Bias} \label{pcdata}
As mentioned, we apply the above ML pipeline to an investigation of pancreatic cancer targeting 3 sampled dataset from the PLCO cohort in order to identify predictive features, assess potential biases in ML analyses, and compare ML modeling algorithm performance.

\subsubsection{PLCO Full Study Population}
The PLCO cohort is derived from the PLCO screening trial, which is a randomized multicenter trial in the United States (Birmingham, AL; Denver, CO; Detroit, MI; Honolulu, HI; Marshfield, WI; Minneapolis, MN; Pittsburgh, PA; Salt Lake City, UT; St. Louis, MO; and Washington, DC) with 152,810 men and women ages 55 to 74 at baseline, that sought to determine the effectiveness of early detection procedures for prostate, lung, colorectal, and ovarian cancers on disease-specific mortality \cite{prorok2000design}. Details of the study methods have been previously described \cite{prorok2000design}. The PLCO screening arm included approximately 77,000 men and women. Study recruitment and randomization began in November 1993 and was completed in July 2001. Data on demographics, health history, diet and other lifestyle factors (intervention arm only), were collected from self-administered questionnaires at baseline. A second self-administered dietary questionnaire was distributed to the intervention and control arms of the trial between 1998 and 2005 to provide additional dietary data \cite{subar2000evaluation}. Cancer cases were identified by self-report in the annual mail-in survey and cohort participants were also linked to cancer registries and the National Death Index. Medical and pathology records were obtained if possible and cancer cases confirmed by study staff \cite{stolzenberg2015circulating}.  In this study, we included incident primary pancreatic adenocarcinomas [International Classification of Diseases, ICD-O-3 code C250-C259 or C25.0-C25.3, C25.7-C25.9] diagnosed between 1994 and 2014.  In total, 800 confirmed pancreatic adenocarcinoma cases were identified across both study arms for this analysis.

\subsubsection{Pancreatic Cancer Case-Control Subsets} \label{panc_data}
We derived three case-control analytic data sets within the PLCO study each defined by different a) case-control selection criteria, b) bias considerations, and c) available features (see Table \ref{tab1}).  The first two datasets include all confirmed pancreatic cases diagnosed in the full cohort (n=800) and only healthy controls with available genotyping data from previous genome-wide association studies (n=4298) \cite{wolpin2014genome,thomas2008multiple,landi2009genome}. Future work will include genotyping data, which is why our control sample is limited to only those 4298 out of over 100,000 participants. Thus, Datasets 1 and 2 introduce selection bias due to control selection.  In particular, control instances included more males (85$\%$) and smokers than would have been expected by chance, which beyond selection bias, can also introduce confounding.  Dataset 2 is very similar to Dataset 1 but adds 14 dietary features (see Section \ref{features}) which include a much higher frequency of missing values in cases, i.e. 19$\%$ in contrast with controls at 2$\%$. Dataset 2 is considered to be the most biased sample in this study as it includes the biases from Dataset 1 along with this missing data. Dataset 3 includes the same features as Dataset 2, but is considered to be the least biased dataset in this study given that it employs a matched case-control design that includes matching on potential confounding factors (i.e. age, gender, race, sex, and data of blood draw), as well as minimizes the the missing data bias in dietary features.  The epidemiological strategy of matching controls to cases decreases confounding and variability related to seasonal variation (i.e. diet), thus improving the precision of the association. However, the third dataset was derived from PLCO studies designed to evaluate effects of biomarkers on pancreatic cancer and included a smaller sample of 328 cases and 652 matched controls that were alive when the matched case was diagnosed \cite{stolzenberg2015circulating}. Matched samples are identified by a unique group ID column in Dataset 3, that is utilized by the ML pipeline to apply matched CV partitioning.  Stratified CV partitioning is used for Datasets 1 and 2. 

\begin{table}[h]
\centering
\begin{tabular}{c c c c l}
\hline
\textbf{Dataset} & \textbf{Cases} & \textbf{Controls} & \textbf{Features} & \textbf{Major Bias Considerations}\\
\hline
Dataset 1 & 800 & 4298 & 24 & Sample selection bias\\ \hline
Dataset 2 & 800 & 4298 & 38 & $+$ Dietary feature missingness\\ \hline
Dataset 3 & 328 & 652 & 38 & Minimal Selection Bias $+$ \\
&&&&Confounding adjustment \\
\hline
\end{tabular}
\caption{Case-control subsets derived from the Full PLCO cohort}
\label{tab1}
\end{table}

\subsubsection{Candidate Risk Factors, i.e. Features} \label{features}
The set of universal risk factors examined in our three target datasets include: (1) established risk factors that are often matched upon and that are either innate (i.e. age, gender, race/ethnicity) or related to study variability (i.e. center, randomization year), (2) established modifiable pancreatic cancer risk factors; i.e. cigarette smoking (yes/no, former smoker, duration/years smoked, pack-years smoked), current body mass index (BMI), diabetes (yes/no), and pancreatic cancer family history (yes/no), (3) other factors that have been variably associated with pancreatic cancer; i.e. education status, BMI at age 20 and age 50, any family history cancer (yes/no), gallbladder and liver disease (yes/no), and (4) risk factors that have not been associated with pancreatic cancer; i.e. marital status, occupation, medication use aspirin (yes/no), ibuprofen (yes/no), daily dose of aspirin and ibuprofen. 

Further, Datasets 2 and 3 include \emph{dietary factors} that have been variably associated with pancreatic cancer including; glycemic index, glycemic load, and total daily intake of carbohydrates, energy, alcohol, fat, folate, red meat, protein, cholesterol, and calcium. In particular, we wanted to determine whether the dietary exposures could improve prediction models. This entire collection of candidate risk factors is based on literature review \cite{stolzenberg2015epidemiology}. 

\section{Results and Discussion for Real World Data}
In this section we present summary results from the application of our proposed ML pipeline to the three sampled pancreatic cancer datasets. The full spectrum of results, figures, statistical analysis, and models from these analyses are available at the second link indicated in Section \ref{mlpipe}.

\subsection{Predictive Performance of ML Models}
Tables \ref{tab2} and \ref{tab3} summarize average balanced accuracy and F1 Score, respectively, for each ML algorithm and pancreatic cancer dataset. Both metrics are suited to evaluate classification performance within imbalanced data, however balanced accuracy equally weights the importance of both classes, while F1 Score focuses on the prediction of the 'case' minority class. Top performing algorithms for each dataset are highlighted in blue, while those whose performance was not significantly worse than the best (p $<$ 0.05 with pairwise Mann-Whitney U-test) are highlighted in yellow. Thus, all highlighted cells are considered to have yielded 'similar' performance for the given dataset. Some high level observations from these tables include: (1) the 'best' performing ML algorithm was different for each dataset, and differed based on the evaluation metric examined, (2) multiple ML algorithms performed similarly well for each dataset, (3) ANN and LCS (i.e. ExSTraCS) with QRF both consistently under-performed other algorithms in this study, (4) LR and RF were the only algorithms to perform 'significantly well' in each dataset of this study; it's possible LR performed so well because the majority of established risk factors for pancreatic cancer were identified in previous studies using logistic regression approaches, and (5) despite LCS never performing 'best', and only sometimes performing 'significantly well', it's performance was never far behind other ML algorithms performing 'significantly well' (in contrast to XGBoost and LGBoost performing best in datsets 1 and 2, but performing poorly in dataset 3).

Regarding the third observation, ANN underperformed despite a fairly extensive hyperparameter sweep that included variations in the depth and width of hidden layers. The application of QRF rule compaction to a trained LCS model \cite{tan2013rapid} was designed to improve model interpretability, which clearly had a negative impact on model performance in this context. Also keep in mind that LCS, with and without QRF was the only algorithm in this study to use default run parameters thus not benefiting from the optimization of a hyperparameter sweep. Anecdotally, an earlier version of this pipeline was implemented and applied to these same datasets, adopting a much simpler 'grid-search-based' hyperparameter sweep. This preliminary analysis yielded significantly worse performance for all other ML algorithms, and suggested LCS performed best in both Datasets 1 and 2. Given that our LCS algorithm, (ExSTraCS) uniquely offers the benefits of model interpretability and the ability to characterize heterogeneous associations, it seems well worth inclusion in other ML analysis pipelines and in our future work that will add genetic variables into this analysis. Despite the computational expense, LCS might strongly benefit from hyperparameter optimization in the future. 

Importantly, the fourth observation above should \emph{not} imply the general superiority of those methods more broadly, as can be clearly seen in contrast with the results on our example SNP dataset in Figure \ref{ml_plots} where LR performs very badly. 

\begin{table}[h]
\centering
\begin{tabular}{l l l l}
\hline
\textbf{ML Algorithm} & \textbf{Dataset 1} & \textbf{Dataset 2} & \textbf{Dataset 3}\\
\hline
Logistic Regression & \cellcolor{yellow!50}0.6795 (0.0328) & \cellcolor{yellow!50}0.6919 (0.0358) & \cellcolor{yellow!50}0.5683 (0.0477) \\ \hline
Decision Tree (DT) & \cellcolor{yellow!50}0.6745 (0.0262) & 0.6637 (0.038) & \cellcolor{yellow!50}0.5551 (0.0475) \\ \hline
Random Forest (RF) & \cellcolor{yellow!50}0.6798 (0.03) & \cellcolor{yellow!50}0.6924 (0.0251) & \cellcolor{yellow!50}0.5582 (0.0468)\\ \hline
Naive Bayes (NB) & 0.6053 (0.024) & 0.622 (0.0249) & \cellcolor{blue!25}0.5729 (0.0395) \\ \hline
XGBoost (XGB) & \cellcolor{blue!25}0.6854 (0.0276) & \cellcolor{yellow!50}0.6948 (0.0267) & 0.5184 (0.0245) \\ \hline
LGBoost (LGB) & \cellcolor{yellow!50}0.6851 (0.0295) & \cellcolor{blue!25}0.7002 (0.0333) & 0.5242 (0.0311) \\ \hline
SVM & \cellcolor{yellow!50}0.6761 (0.0218) & 0.591 (0.0436) & \cellcolor{yellow!50}0.5669 (0.0563) \\ \hline
ANN & 0.5824 (0.0301) & 0.58 (0.0145) & 0.5468 (0.0162) \\ \hline
LCS & 0.6668 (0.0191) & 0.6684 (0.0318) & \cellcolor{yellow!50}0.5536 (0.0342) \\ \hline
LCS with QRF & 0.5579 (0.0164) & 0.5745 (0.0203) & 0.5181 (0.0215) \\ \hline
\hline
\end{tabular}
\caption{Balanced accuracy (with standard deviation) averaged over 10-fold CV for each dataset and ML algorithm. Best performing algorithms are highlighted in blue, while those that did not perform significantly worse (p $<$ 0.05) are highlighted in yellow.}
\label{tab2}
\end{table}

\begin{table}[h]
\centering
\begin{tabular}{l l l l}
\hline
\textbf{ML Algorithm} & \textbf{Dataset 1} & \textbf{Dataset 2} & \textbf{Dataset 3}\\
\hline
Logistic Regression & \cellcolor{yellow!50}0.4221 (0.042) & \cellcolor{yellow!50}0.4303 (0.0417) & \cellcolor{yellow!50}0.3882 (0.0725) \\ \hline
Decision Tree (DT)& \cellcolor{yellow!50}0.4183 (0.0352) & 0.3922 (0.0423) & 0.3231 (0.1397) \\ \hline
Random Forest (RF)& \cellcolor{yellow!50}0.4272 (0.04) & \cellcolor{blue!25}0.4477 (0.0321) & \cellcolor{yellow!50}0.3732 (0.0922)\\ \hline
Naive Bayes (NB) & 0.3383 (0.0474) & 0.3644 (0.044) & \cellcolor{yellow!50}0.4051 (0.0612)\\ \hline
XGBoost (XGB)& \cellcolor{blue!25}0.4317 (0.0363) & \cellcolor{yellow!50}0.438 (0.0325) & 0.3087 (0.0571) \\ \hline
LGBoost (LGB) & \cellcolor{yellow!50}0.4311 (0.0376) & \cellcolor{yellow!50}0.4446 (0.0389) & 0.3401 (0.0585)\\ \hline
SVM & \cellcolor{yellow!50}0.4231 (0.0298) & 0.3043 (0.0663) & \cellcolor{blue!25}0.4436 (0.0739) \\ \hline
ANN & 0.2908 (0.0649) & 0.2866 (0.0297) & 0.2833 (0.0605)\\ \hline
LCS & \cellcolor{yellow!50}0.4215 (0.027) & \cellcolor{yellow!50}0.4211 (0.0441) & 0.3713 (0.0653) \\ \hline
LCS with QRF & 0.2188 (0.0486) & 0.2642 (0.0517) & 0.1312 (0.0655) \\ \hline
\hline
\end{tabular}
\caption{F1 Score (with standard deviation) averaged over 10-fold CV for each dataset and ML algorithm. Best performing algorithms are highlighted in blue, while those that did not perform significantly worse (p $<$ 0.05) are highlighted in yellow.}
\label{tab3}
\end{table}

\begin{figure}[h]
\centering\includegraphics[width=1.0\linewidth]{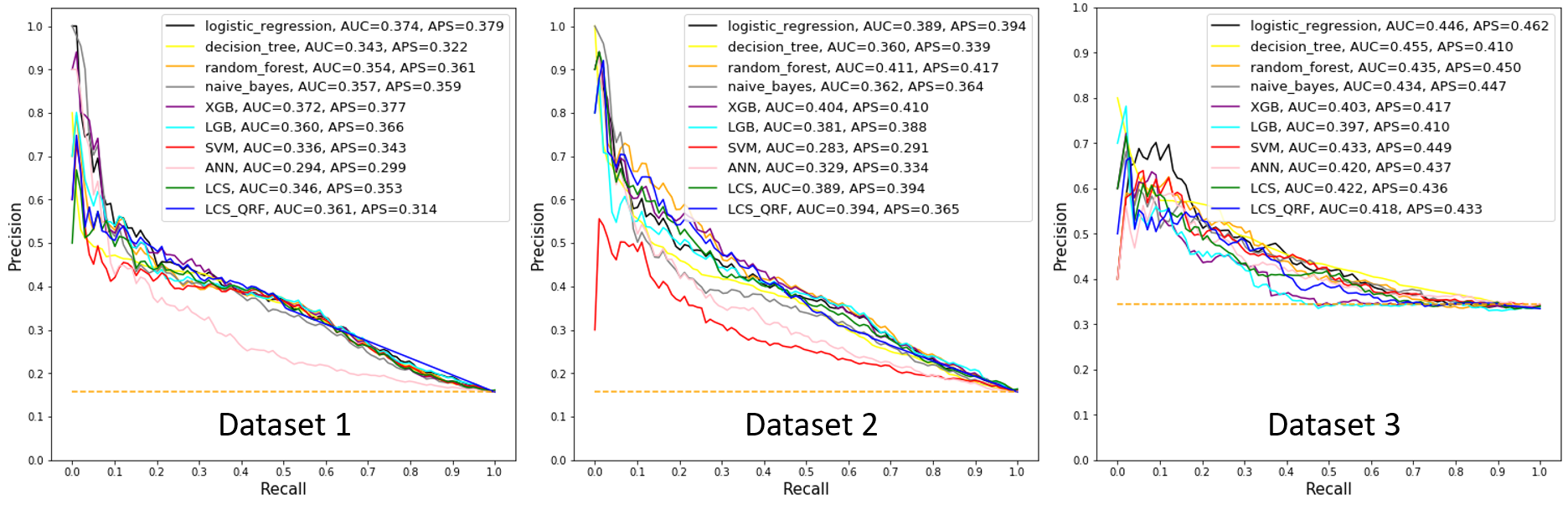}
\caption{PRC plots summarizing ML performance of the three pancreatic cancer datasts with class imbalance. PRC AUC and average precision score (APS) included for each.}
\label{prc_pc}
\end{figure}

Given the presence of class imbalance, Figure \ref{prc_pc} compares ML performance on the same three datasets with precision/recall curve plots. This gives us an additional perspective on performance. Again, depending on the metric you focus on, the 'best' performing ML algorithm is inconsistent within and between datasets, as should be expected. 

In comparing 'best' performance between datasets in Tables \ref{tab2} and \ref{tab3}, and Figure \ref{prc_pc}, we observe the better performances in Dataset 2 (with a larger sample size, greater class imbalance, all features, and the greatest opportunity for bias), followed closely by Dataset 1 (same as 2 but without the dietary features), with Dataset 3 (with the smallest sample size, less class imbalance, all features, and considered to be least biased) performing least well. In general, the best performing algorithms improved performance with the addition of dietary features (Dataset 1 vs. 2) by some small degree (e.g. XGBoost in Table \ref{tab2}, from 0.6854 to 0.6948), however this was not true for all algorithms (e.g. DT, SVM, and ANN). While suggestive, it is ultimately inconclusive whether dietary features improve model performance. Variations may reflect different ML algorithms sensitivity to potential bias introduced by missing value imputation of the dietary variables, or differences in the ability of each ML algorithm to detect the underlying patterns in these datasets.

\subsection{ML Feature Importance}
\begin{figure}[h]
\centering\includegraphics[width=1\linewidth]{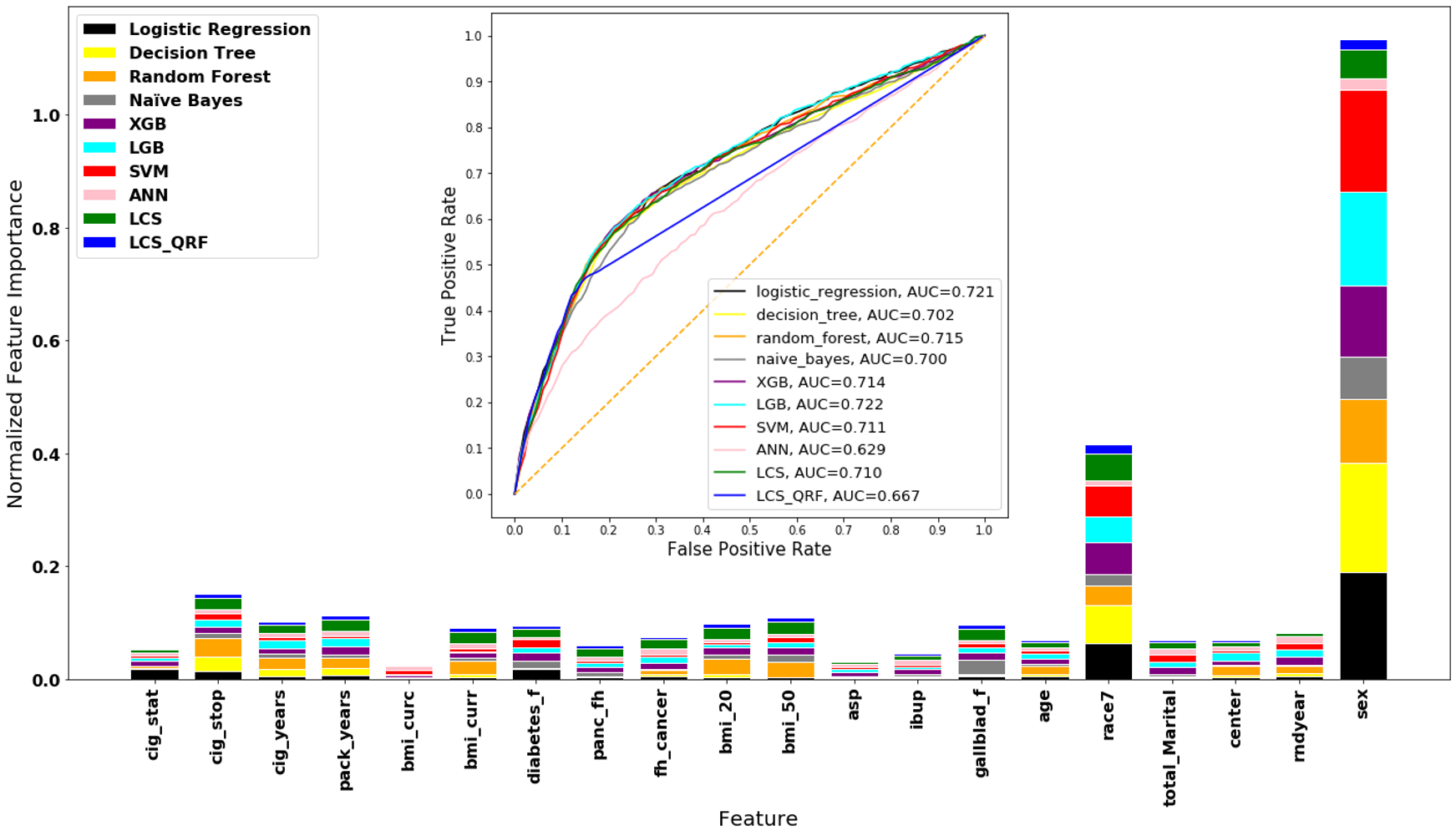}
\caption{Dataset 1 ML feature importance as a CFIBP. The corresponding ROC plot is embedded in this figure in contrast with the first PRC plot in Figure \ref{prc_pc}.}
\label{fi_pc_roc_1}
\end{figure}

\begin{figure}[h]
\centering\includegraphics[width=1\linewidth]{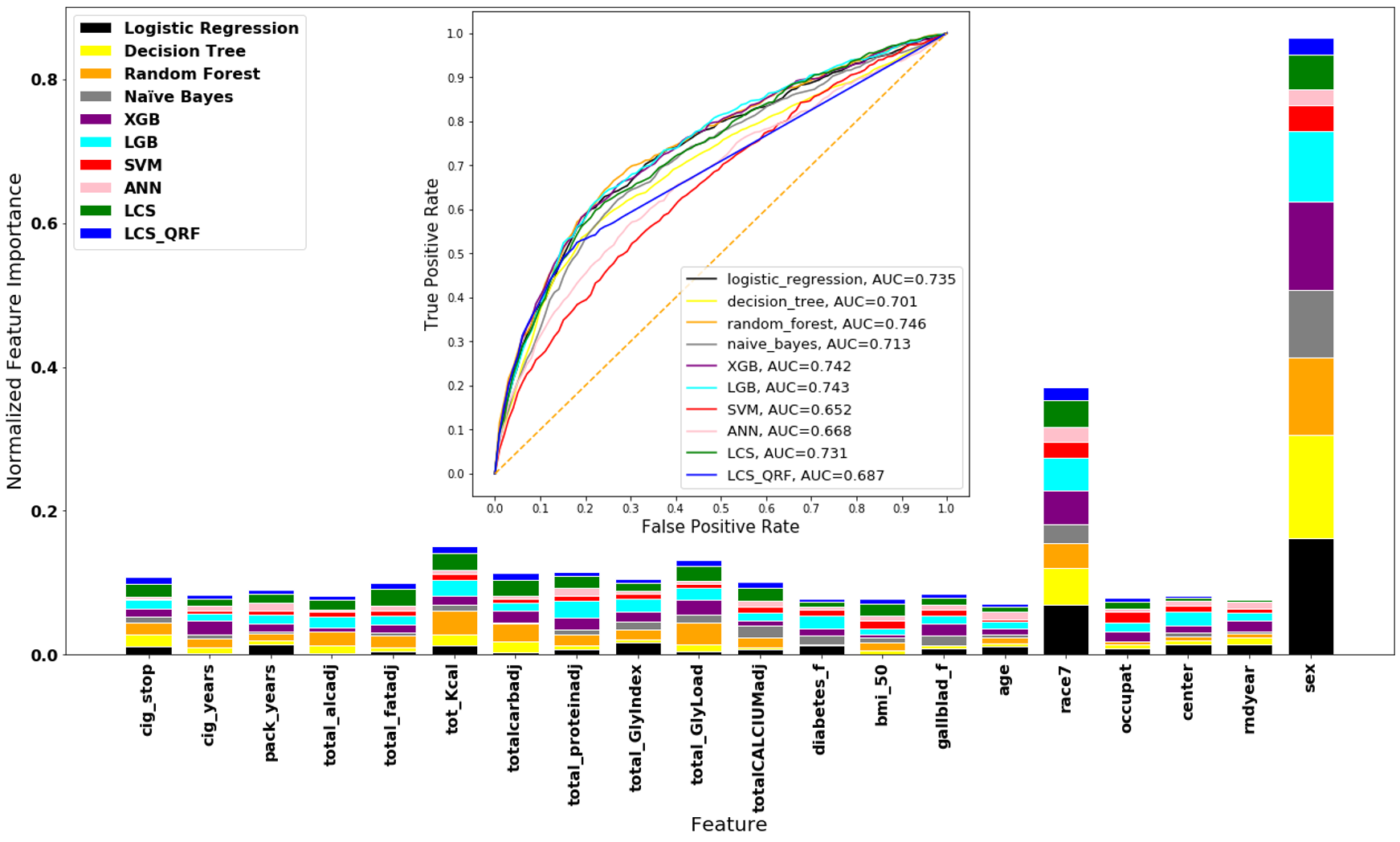}
\caption{Dataset 2 ML feature importance as a CFIBP. The corresponding ROC plot is embedded in this figure in contrast with the second PRC plot in Figure \ref{prc_pc}.}
\label{fi_pc_roc_2}
\end{figure}

\begin{figure}[h]
\centering\includegraphics[width=1\linewidth]{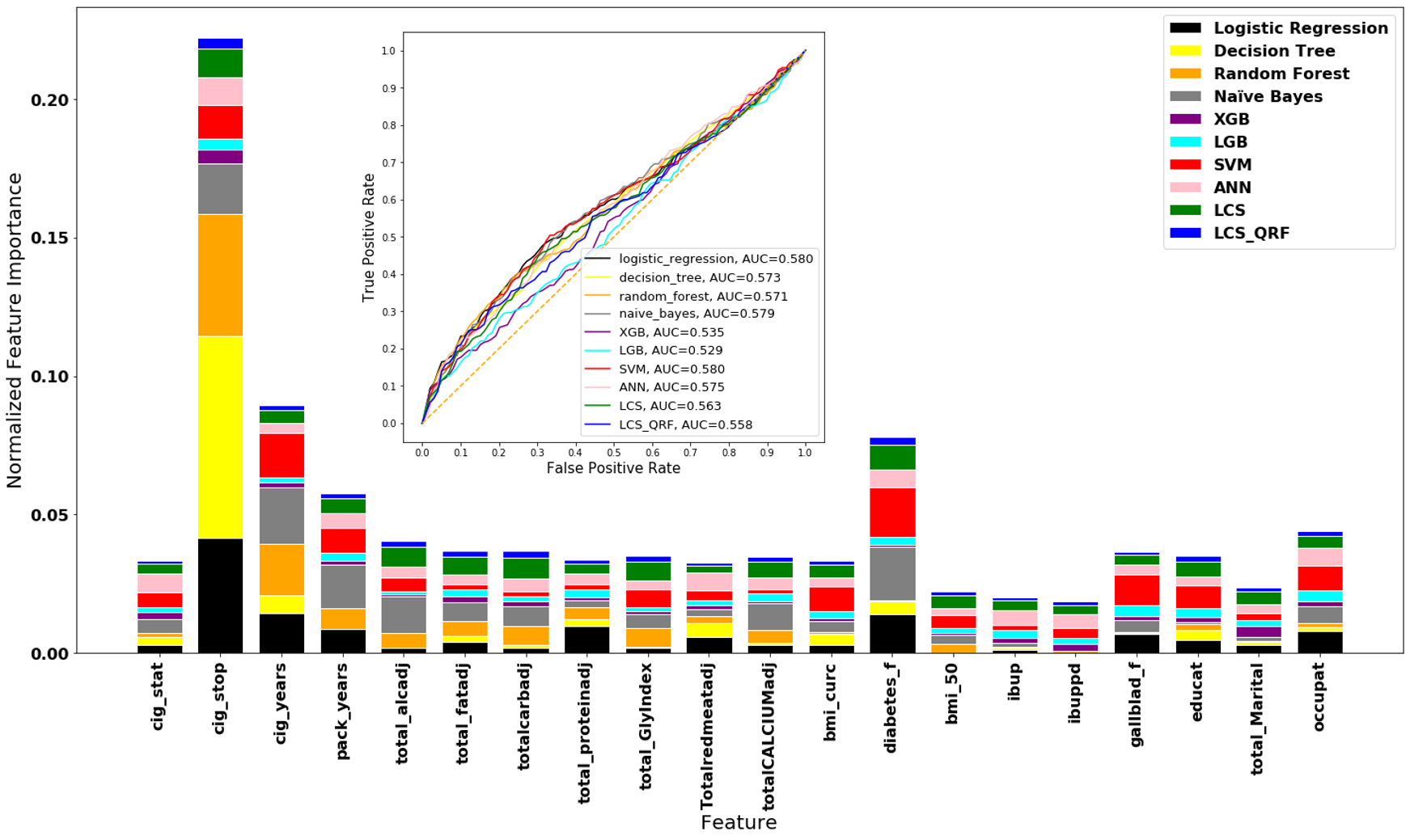}
\caption{Dataset 3 ML feature importance as a CFIBP. The corresponding ROC plot is embedded in this figure in contrast with the third PRC plot in Figure \ref{prc_pc}.}
\label{fi_pc_roc_3}
\end{figure}

Figures \ref{fi_pc_roc_1}, \ref{fi_pc_roc_2}, and \ref{fi_pc_roc_3} give respective CFIBPs illustrating compound average ML algorithm feature importance for Datasets 1-3. Each applies normalization, sum division, and algorithm performance weighting (via balanced accuracy) as in Figure \ref{composite_fi_frac_weight_sim}. Only the top 20 features are included in each.

One of the clearest observations from Figures \ref{fi_pc_roc_1} and \ref{fi_pc_roc_2} is that the predictive performance of most of these algorithms in Datasets 1 and 2 is primarily driven by sex and race, supporting their role as strong covariates and known confounders. However, their importance vanishes in Figure \ref{fi_pc_roc_3}, where these covariates are controlled for by case/control matching. Another expected observation is whenever both the original (continuous) and stratified (categorical) version of a feature is included in the dataset, i.e. ‘age at study entry’ (age) vs. ‘age level’ (agelevel) and ‘BMI’ (bmi
$\_$curc) vs. 'BMI categorical' (bmi$\_$curc), the categorical version of the variable yields a lower feature importance. This reflects the information loss that takes place during variable stratification.

In Figure \ref{fi_pc_roc_1}, focusing on Dataset 1, we note that specific risk factors such as 'years since quit smoking' (cig$\_$stop), pack-years of smoking (pack$\_$years), the three BMI variables, i.e. 'current BMI' (bmi$\_$curr), 'BMI at age 20' (BMI$\_$20), and 'BMI at age 50' (bmi$\_$50), and 'history of diabetes' (diabetes$\_$f) and 'history of gallbladder' (gallblad$\_$f) have the next highest compound feature importance over all ML algorithms. Interestingly, the 'age' covariate, a known risk factor and confounder had a relatively low effect in this dataset. 

In Figure \ref{fi_pc_roc_2}, focusing on Dataset 2, we note that beyond sex and race, a number of the dietary features yielded the strongest compound feature importance including 'total Kcal' or daily calorie intake (tot$\_$Kcal), and 'total glyLoad' or glycemic load (total$\_$GlyLoad). Consistent with Dataset 1, cig$\_$stop, pack$\_$years, bmi$\_$50, diabetes$\_$f, and gallblad$\_$f also stood out among the top 20 out of 38 total features. 

In Figure \ref{fi_pc_roc_3}, focusing on Dataset 3, cig$\_$stop stands out the most, followed by cig$\_$years, diabetes$\_$f, and pack$\_$years.  Similar to Dataset 2, dietary features yielded strong feature importance scores, though total alcohol consumption demonstrated the highest feature importance among dietary factors.  However, compared to Dataset 2, total Kcal and total glyLoad (glycemic load) were not identified as important features in the matched case-control design.  Give that all algorithms struggled to perform well on this dataset, it is perhaps more informative to focus on the feature importance scores of the best performing algorithm including LR, RF, NB, and SVM. Further it is useful to point out that in each of these figures relative feature importance is not always consistent from one ML algorithm to the next (e.g. in Figure \ref{fi_pc_roc_3}, cig$\_$stop is most important for DT, but cig$\_$years is most important for NB). In some cases this may be the result of different feature importance estimation strategies being used by some of the ML algorithms, but in others it reflects differences in how distinct ML algorithm representations and learning schemes function on the same data.

In summary, we observe that across algorithms which successfully train a predictive model (i.e. balanced accuracy $>$ 0.55), the “important” features identified by those models were largely consistent across the case-control sets in which those variables were present (e.g. smoking variables). This reproducibility would suggest that these features or risk factors are more likely to be true predictors despite the presence of known biases. However, the relative importance (i.e. ranking) of variable features did vary across case-control sets by machine learner. 

These findings suggest that dietary factors warrant further investigation in pancreatic cancer risk models and also suggest that while most ML studies evaluate a single definition of case or control status, it seems advantageous for discovery and interpretation to construct multiple case-control datasets and to test multiple MLs in cohort studies as a strategy to examine consistency of results (and target inconsistencies for follow up/interpretation). 

Specifically, we suggest the utilization of ML analysis pipelines, like the one we developed here, and the generation of different case-control sets from the master data source, based on expert knowledge of potential biases, to assess reliability and consistency.  Each data resource, whether it is a cohort or EHRs, is susceptible to biases that arise from the underlying data structure of that particular sample.  Thus, coupling empiric prediction approaches with expert knowledge that attempts to address potential biases due to study sample selection could make the choice of the machine learner and the validity of subsequent findings (particularly in discovery studies), less of a “black box” and more acceptable in population studies. If this option is computationally restrictive we suggest utilizing the complete dataset, with the maximal number of potentially predictive variables, as well as the maximum number of cases and controls to improve the chance of discovery and maximizing predictive ML performance.  However, in these instances, additional adjustments for bias, including residual adjustment for confounders and leave-one-out feature analyses should be considered \cite{brookhart2010confounding}.

\section{Conclusions and Ongoing Work}
In this study we designed, implemented, and applied a rigorous ML analysis pipeline for binary classification that emphasizes ML best practices, transparency, reproducibility, automation, justification, and the ability to detect complex associations. This pipeline may be applied as-is or expanded upon for future applications. This pipeline was organized as 4 basic stages including: 1) data preprocessing and feature transformation, 2) feature importance and feature selection with collective FS, 3) ML modeling with 9 established algorithms, and 4) post-analysis including compound feature importance bar plots (CFIBPs). Through a simulated genetic example with complex associations and followup application to 3 different case-control pancreatic cancer datasets we demonstrate the efficacy and utility of this pipeline as a proposed rigorous standard for biomedical ML. We expect this automated pipeline to offer a more transparent and customizable alternative to AutoML, that will serve to educate and encourage less biased and more rigorous applications of ML to biomedical analyses in the future. 

Included in this pipeline is our own ExSTraCS algorithm, rule-based 'LCS' machine learner that had previously been demonstrated to proficiently detect and characterize epistatic and heterogeneous patterns of association. We utilize this pipeline to compare the performance of ExSTraCS to other well-known ML algorithms, despite being at the disadvantage of using default run parameters rather than optimizing via a hyperparameter sweep. While performing best on the simulated genetic example, in this particular study of pancreatic cancer did not always achieve the best predictive performance, but still performed well in the greatest variety of situations (i.e. dataset and evaluation metric). Given LCS's interpretability advantages and ability to detect complex associations, including genetic heterogeneity, we submit that ExSTraCS and LCS algorithms are valuable to include for comparison within future ML analysis pipelines. While most methods in this pipeline are scikit-learn compatible, our ExSTraCS algorithm is a stand-alone Python implementation. This illustrates how a user might adapt this pipeline to compare their own novel ML algorithms that may or may not be scikit-learn compatible. 

Analyses on three target pancreatic cancer datasets a) demonstrated the impact that bias may have on ML and feature selection, b) confirmed known risk factors, c) confirmed the efficacy of case/control matching of confounding variables to eliminate their effect, d) suggested that dietary variables may contribute a small predictive advantage when included in ML models, though which dietary factors are most informative were not consistent, and e) identified subsets of ML algorithms that performed well on each dataset. 

Ongoing work will seek to expand these efforts in a number of ways: a) add other appropriate ML algorithms for comparison, e.g. nearest neighbors classifier \cite{hu2016distance} and other scikit-learn compatible LCS algorithms being developed, b) add other appropriate FS algorithms to expand the capabilities of collective feature selection, c) expand FS with appropriate wrapper algorithms to improve scalability of this pipeline to datasets with $>$ 10,000 features, e) utilize this pipeline as a blueprint to construct similar pipelines for multi-class and quantitative (i.e. regression) outcomes, f) compartmentalize and set up parallelization of this pipeline outside of a Jupyter notebook so that it can be more efficiently run on larger-scale datasets, g) expand these pancreatic cancer analysis to include genetic variables, h) utilize the unique capabilities of ExStraCS to characterize potential heterogeneous associations, and i) expand the model interpretation capabilities of this pipeline beyond feature importance assessment wherever possible. 

\section*{Acknowledgements}
Special thanks to Drs. Patryk Orzechowski and Trang Le for their feedback in the development of this pipeline. This work is supported by a grant from the Department of Defense to SML (W81XWH-17-1-0276) and NIH grants to JHM (LM010098, LM012601, AI116794).

\bibliographystyle{abbrv}


\end{document}